\documentclass[11pt,a4paper]{article}
\usepackage{authblk}
\usepackage[hyperref]{emnlp2020}
\usepackage{times}
\usepackage{latexsym}

\usepackage{tabularx}
\usepackage{makecell}
\usepackage{booktabs}
\usepackage{amsmath}
\usepackage{soul}
\usepackage{graphicx}

\usepackage{pifont}
\definecolor{rred}{HTML}{C0504D}
\definecolor{ggreen}{HTML}{9BBB59}
\newcommand{\cmark}{\textcolor{ggreen}{\ding{51}}}
\newcommand{\xmark}{\textcolor{rred}{\ding{55}}}

\usepackage{microtype}

\aclfinalcopy 
\def\aclpaperid{2114} 

\title{Detecting Word Sense Disambiguation Biases in Machine Translation\\for Model-Agnostic Adversarial Attacks}

\author[1]{Denis Emelin}
\author[1, 2]{Ivan Titov}
\author[3, 1]{Rico Sennrich}

\affil[1]{University of Edinburgh, Scotland}
\affil[2]{University of Amsterdam, Netherlands}
\affil[3]{University of Zurich, Switzerland}
\affil[ ]{\texttt {D.Emelin@sms.ed.ac.uk}} 
\affil[ ]{\texttt {ititov@inf.ed.ac.uk sennrich@cl.uzh.ch}}

\date{}

\begin{document}
\maketitle
\begin{abstract}
Word sense disambiguation is a well-known source of translation errors in NMT. We posit that some of the incorrect disambiguation choices are due to models' over-reliance on dataset artifacts found in training data, specifically superficial word co-occurrences, rather than a deeper understanding of the source text. We introduce a method for the prediction of disambiguation errors based on statistical data properties, demonstrating its effectiveness across several domains and model types. Moreover, we develop a simple adversarial attack strategy that minimally perturbs sentences in order to elicit disambiguation errors to further probe the robustness of translation models. Our findings indicate that disambiguation robustness varies substantially between domains and that different models trained on the same data are vulnerable to different attacks.\footnote{Experimental codebase available at \url{http://github.com/demelin/detecting_wsd_biases_for_nmt}}
\end{abstract}

\section{Introduction}
Consider the sentence \textit{John met his wife in the hot spring of 1988}. In this context, the polysemous term \textit{spring} unambiguously refers to the season of a specific year. Its appropriate translation into German would therefore be \textit{Frühling} (the season), rather than one of its alternative senses, such as \textit{Quelle} (the source of a stream). To contemporary machine translation systems, however, this sentence presents a non-trivial challenge, with Google Translate (GT) producing the following translation: \textit{John traf seine Frau in der heißen \underline{Quelle} von 1988.}

Prior studies have indicated that neural machine translation (NMT) models rely heavily on source sentence information when resolving lexical ambiguity \cite{tang2019encoders}. This suggests that the combined source contexts in which a specific sense of an ambiguous term occurs in the training data greatly inform the models' disambiguation decisions. Thus, a stronger correlation between the English collocation \textit{hot spring} and the German translation \textit{Quelle}, as opposed to \textit{Frühling}, in the training corpus may explain this disambiguation error. Indeed, \textit{John met his wife in the spring of 1988} is translated correctly by GT.

We propose that our motivating example is representative of a systematic pathology NMT systems have yet to overcome when performing word sense disambiguation (WSD). Specifically, we hypothesize that translation models learn to disproportionately rely on lexical correlations observed in the training data when resolving word sense ambiguity. As a result, disambiguation errors are likely to arise when an ambiguous word co-occurs with words that are strongly correlated in the training corpus with a sense that differs from the reference.

To test our hypothesis, we evaluate whether dataset artifacts are predictive of disambiguation decisions made in NMT. First, given an ambiguous term, we define a strategy for quantifying how much its context biases NMT models towards its different target senses, based on statistical patterns in the training data. We validate our approach by examining correlations between this bias measure and WSD errors made by baseline models. Furthermore, we investigate whether such biases can be exploited for the generation of minimally-perturbed adversarial samples that trigger disambiguation errors. Our method does not require access to gradient information nor the score distribution of the decoder, generates samples that do not significantly diverge from the training domain, and comes with a clearly-defined notion of attack success and failure.

The main contributions of this study are:
\begin{enumerate}
\item We present evidence for the over-reliance of NMT systems on inappropriate lexical correlations when translating polysemous words.
\item We propose a method for quantifying WSD biases that can predict disambiguation errors.
\item We leverage data artifacts for the creation of adversarial samples that facilitate WSD errors.
\end{enumerate}

\section{Can WSD errors be predicted?}
\label{sec:can_wsd}

To evaluate whether WSD errors can be effectively predicted, we first propose a method for measuring the bias of sentence contexts towards different senses of polysemous words, based on lexical co-occurrence statistics of the training distribution. We restrict our investigation to English$\to$German, although the presented findings can be assumed to be language-agnostic. 
To bolster the robustness of our results, we conduct experiments in two domains - movie subtitles characterized by casual language use, and the more formal news domain. For the former, we use the OpenSubtitles2018 (OS18) \cite{lison2019open} corpus\footnote{\url{http://opus.nlpl.eu}}, whereas the latter is represented by data made available for the news translation task of the Fourth Conference on Machine Translation (WMT19)\footnote{\url{http://statmt.org/wmt19}} \cite{barrault2019findings}. Appendix \ref{appendix:data} reports detailed corpus statistics.

\subsection{Quantifying disambiguation biases}
\label{sec:quantifying}

An evaluation of cross-lingual WSD errors presupposes the availability of certain resources, including a list of ambiguous words, a lexicon containing their possible translations, and a set of parallel sentences serving as a disambiguation benchmark. 

\subsection*{Resource collection}

Since lexical ambiguity is a pervasive feature of natural language, we limit our study to homographs - polysemous words that share their written form but have multiple, unrelated meanings. We further restrict the set of English homographs to nouns that are translated as distinct German nouns, so as to confidently identify disambiguation errors, while minimizing the models' ability to disambiguate based on syntactic cues.
English homographs are collected from web resources\footnote{\url{http://7esl.com/homographs}\\\url{http://en.wikipedia.org/wiki/List_of_English_homographs}}, excluding those that do not satisfy the above criteria. Refer to appendix \ref{appendix:homographs} for the full homograph list.

We next compile a parallel lexicon of homograph translations, prioritizing a high coverage of all possible senses. Similar to \cite{raganato2019mucow}, we obtain sense-specific translations from cross-lingual BabelNet \cite{navigli2010babelnet} synsets. Since BabelNet entries vary in their granularity, we iteratively merge related synsets as long as they  have at least three German translations in common or share at least one definition.\footnote{A manual inspection found the clusters to be meaningful.} This leaves us with multiple sense clusters of semantically related German translations per homograph. To further improve the quality of the lexicon, we manually clean and extend each homograph entry to address the noise inherent in BabelNet and its incomplete coverage.\footnote{The lexicon is released as part of our experimental code: \url{http://github.com/demelin/detecting_wsd_biases_for_nmt}.} Appendix \ref{appendix:clusters} provides examples of the final sense clusters.

In order to identify sentence contexts specific to each homograph sense, parallel sentences containing known homographs are extracted from the training corpora in both domains. We lemmatize homographs, their senses, and all sentence pairs using spaCy \cite{honnibal2017spacy} to improve the extraction recall. Homographs are further required to be aligned with their target senses according to alignments learned with fast\_align \citep{dyer2013simple}. Each extracted pair is assigned to one homograph sense cluster based on its reference homograph translation. Pairs containing homograph senses assigned to multiple clusters are ignored, as disambiguation errors are impossible to detect in such cases.

\subsection*{Bias measures}

It can be reasonably assumed that context words co-occurring with homographs in a corpus of natural text are more strongly associated with some of their senses than others. Words that are strongly correlated with a specific sense may therefore bias models towards the corresponding translation at test time. We refer to any source word that co-occurs with a homograph as an \textit{attractor} associated with the sense cluster of the homograph's translation. Similarly, we denote the degree of an attractor's association with a sense cluster as its \textit{disambiguation bias} towards that cluster. Table \ref{tab:attractors} lists the most frequent attractors identified for the different senses of the homograph \textit{spring} in the OS18 training set. 
\begin{table}[!h]
\centering
\begin{tabular}{c  c  c}
\toprule
\textit{season} & \textit{water source} & \textit{device} \\
\midrule
summer & hot & like  \\
winter & water & back \\
come & find & thing \\
\bottomrule
\end{tabular}
\caption{Examples of attractors for \textit{spring}.}
\label{tab:attractors}
\end{table}

Intuitively, if an NMT model disproportionately relies on simple surface-level correlations when resolving lexical ambiguity, it is more likely to make WSD errors when translating sentences that contain strong attractors towards a wrong sense. To test this, we collect attractors from the extracted parallel sentences, quantifying their disambiguation bias (DB) using two metrics: Raw co-occurrence frequency (FREQ) and positive point-wise mutual information (PPMI) between attractors and homograph senses. FREQ is defined in Eqn.\ref{eq:1}, while Eqn.\ref{eq:2} describes PPMI, with $w \in V$ denoting an attractor term in the source vocabulary\footnote{We consider any word that co-occurs with a homograph in the training corpus as an attractor of the homograph's specific sense cluster, except for the homograph itself which is not regarded as an attractor for any of its known sense clusters.}, and $sc \in SC$ denoting a sense cluster in the set of sense clusters assigned to a homograph. For PPMI, $P(w_i, sc_j)$, $P(w_i)$, and $P(sc_j)$ are estimated via relative frequencies of (co-)occurrences in training pairs.
\begin{align}
FREQ(w_i, sc_j) &= Count(w_i, sc_j)\label{eq:1}\\
PPMI(w_i, sc_j) &= max(\frac{P(w_i,sc_j)}{P(w_i) P(sc_j)}, 0)\label{eq:2}
\end{align}
The disambiguation bias associated with the entire context of a homograph is obtained by averaging sense-specific bias values $DB(w_i, sc_j)$ of all attractors in the source sentence $S=\{w_1, w_2, ..., w_{|S|}\}$, as formalized in Eqn.\ref{eq:3}. Context words that are not known attractors of $sc_j$ are assigned a disambiguation bias value of 0.
\begin{equation}\label{eq:3}
DB(S, sc_j) = \frac{1}{|S|} \sum_{i=1}^{|S|} DB(w_i, sc_j)
\end{equation}
As a result, sentences containing a greater number of strong attractors are assigned a higher bias score. 

\subsection{Probing NMT models}
\label{sec:probing}

To evaluate the extent to which sentence-level disambiguation bias is predictive of WSD errors made by NMT systems, we train baseline translation models for both domains. The baselines include Transformer \cite{vaswani2017attention}, LSTM \cite{luong2015effective}, and convolutional Seq-to-Seq (ConvS2S) \cite{gehring2017convolutional} models of comparable size. Appendix \ref{appendix:models} details the training regime and hyper-parameter choices. SacreBLEU \cite{post2018call} scores reported in Table \ref{tab:baselines} indicate that all models are reasonably competent.
\begin{table}[!h]
\centering
\resizebox{7.5cm}{!}{%
\begin{tabular}{l c cc}
\toprule
& \multicolumn{1}{c}{} & \multicolumn{2}{c}{\textbf{WMT}} \\
\cmidrule(lr){3-4}
\textbf{Architecture} & \textbf{OS18 test} & \textbf{test 2014} & \textbf{test 2019}  \\
\midrule
Transformer & 29.7 & 27.5 & 38.2  \\
\midrule
LSTM & 27.7 & 24.1 & 34.3 \\
\midrule
ConvS2S & 27.7 & 23.5 & 32.5 \\
\bottomrule
\end{tabular}}
\caption{EN-DE translation performance (BLEU).}
\label{tab:baselines}
\end{table}

Test sets for WSD error prediction are constructed by extracting parallel sentences from held-out, development, and test data (see appendix \ref{appendix:data} for details). The process is identical to that described in section \ref{sec:quantifying}, with the added exclusion of source sentences shorter than 10 tokens, as they may not provide enough context. For each source sentence, disambiguation bias values are computed according to equation \ref{eq:3}, with $sc_j$ corresponding to either the correct sense cluster (DB\textsubscript{\cmark}) or the incorrect sense cluster with the strongest bias (DB\textsubscript{\xmark}). Additionally, we consider the difference DB\textsubscript{DIFF} between DB\textsubscript{\xmark} and DB\textsubscript{\cmark} which can be interpreted as the overall statistical bias in a source sentence towards an incorrect homograph translation. All bias scores are computed either using FREQ or PPMI.

\begin{table*}[!t]
\centering
\resizebox{14.5cm}{!}{%
\begin{tabular}{l c c c c c c c}
\toprule
\textbf{Model} & \textbf{FREQ\textsubscript{\cmark}} & \textbf{PPMI\textsubscript{\cmark}} & \textbf{FREQ\textsubscript{\xmark}} & \textbf{PPMI\textsubscript{\xmark}} & \textbf{FREQ\textsubscript{DIFF}} & \textbf{PPMI\textsubscript{DIFF}} & \textbf{Length}\\
\midrule
\midrule
OS18 Transformer & -0.532 & -0.578 & 0.327 & 0.474 & \textbf{0.708} & 0.674 & 0.018\\ 
\midrule
OS18 LSTM & -0.468 & -0.504 & 0.386 & 0.486 & \textbf{0.690} & 0.630 & 0.008\\
\midrule
OS18 ConvS2S & -0.477 & -0.514 & 0.391 & 0.492 & \textbf{0.723} & 0.658 & 0.021\\
\midrule
\midrule
WMT19 Transformer & -0.610 & -0.668 & 0.415 & 0.579 & \textbf{0.687} & 0.677 & -0.004\\ 
\midrule
WMT19 LSTM & -0.661 & -0.698 & 0.376 & 0.574 & \textbf{0.725} & 0.708 & -0.009\\
\midrule
WMT19 ConvS2S & -0.648 & -0.678 & 0.408 & 0.599 & \textbf{0.731} & 0.710 & 0.000\\
\bottomrule
\end{tabular}}
\caption{Rank biserial correlation between disambiguation bias measures and lexical disambiguation errors.}
\label{tab:seed_corr}
\end{table*}

We examine correlations between the proposed bias measures and WSD errors produced by the in-domain baseline models. Translations are considered to contain WSD errors if the target homograph sense does not belong to the same sense cluster as its reference translation. We check this by looking up target words aligned with source homographs according to fast\_align. To estimate correlation strength we employ the ranked biserial correlation (RBC) metric\footnote{We additionally used the non-parametric generalization of the Common Language Effect Size \cite{ruscio2008probability} for correlation size estimation, but couldn't detect any advantages over RBC in our experimental setting.} \cite{cureton1956rank} and measure statistical significance using the Mann-Whitney U (MWU) test \cite{mann1947test}. 

In order to compute the RBC values, test sentences are divided into two groups - one containing correctly translated source sentences and another comprised of source sentences with incorrect homograph translations. Next, all possible pairs are constructed between the two groups, pairing together each source sentence from one group with all source sentences from the other. Finally, the proportion of pairs $f$ where the DB score of the incorrectly translated sentence is greater than that of the correctly translated sentence is computed, as well as the proportion of pairs $u$ where the opposite relation holds. The RBC value is then obtained according to Eqn.\ref{eq:4}.  
\begin{equation}\label{eq:4}
RBC = f - u
\end{equation}
Statistical significance, on the other hand, is estimated by ranking all sentences in the test set according to their DB score in ascending order while resolving ties, and computing the U-value according to Eqn.\ref{eq:5}-\ref{eq:7}, where $R_1$ denotes to the sum of ranks of sentences with incorrectly translated homographs and $n_1$ their total count, while $R_2$ denotes the sum of ranks of correctly translated sentences and $n_2$ their respective total count.  
\begin{align}
U &= min(U_1, U_2)\label{eq:5}\\
U_1 &= R_1 - \frac{n_1 (n_1 + 1)}{2}\label{eq:6}\\ 
U_2 &= R_2 - \frac{n_2 (n_2 + 1)}{2}\label{eq:7}
\end{align}
To obtain the p-values, U-values are subjected to tie correction and normal approximation.\footnote{We use Python implementations of RBC and MWU provided by the pingouin library \cite{Vallat2018}.}

Table \ref{tab:seed_corr} summarizes the results\footnote{Positive values denote a positive correlation between bias measures and the presence of disambiguation errors in model translations, whereas negative values denote negative correlations. The magnitude of the values, meanwhile, indicates the correlations' effect size.}, including correlation estimates between WSD errors and source sentence length, as a proxy for disambiguation context size. Statistically significant correlations are discovered for all bias estimates based on attractors (p $<$ 1e-5, two-sided). Moreover, the observed correlations exhibit a strong effect size \cite{mcgrath2006effect}. See appendix \ref{appendix:effect_size} for the model-specific effect size interpretation thresholds. For all models and domains the strongest correlations are observed for DB\textsubscript{DIFF} derived from simple co-occurrence counts.

\subsection*{Challenge set evaluation}
\begin{figure}[!h]
\centering
\includegraphics[width=8cm]{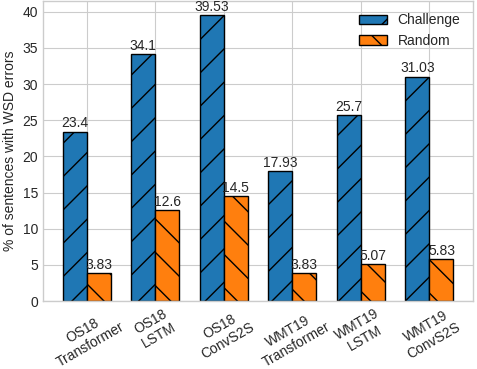}
\caption{WSD errors in subsampled challenge sets.}
\label{fig:seed_challenge}
\end{figure}

To establish the predictive power of the uncovered correlations, a challenge set of 3000 test pairs with the highest FREQ\textsubscript{DIFF} score is subsampled from the full WSD test pair pool in both domains. In addition, we create secondary sets of equal size by randomly selecting pairs from each pool. As Figure \ref{fig:seed_challenge} illustrates, our translation models exhibit a significantly higher WSD error rate - by a factor of up to \textbf{6.1} - on the challenge sets as compared to the randomly chosen pairs. While WSD performance is up to 96\% on randomly chosen sentences, performance drops to 77--82\% for the best-performing model (Transformer). This suggests that lexical association artifacts, from which the proposed disambiguation bias measure is derived, can be an effective predictor of lexical ambiguity resolution errors across model architectures and domains. 

The observed efficacy of attractor co-occurrence counts for WSD error prediction may be partially due to sense frequency effects, since more frequent senses occur in more sentence pairs, yielding more frequent attractors. NMT models are known to underperform on low-frequency senses of ambiguous terms \cite{gonzales2017improving}, prompting us to investigate if disambiguation biases capture the same information. For this purpose, another challenge set of 3000 pairs is constructed by prioritizing pairs assigned to the rarest among each homograph's sense sets. We find that the new challenge set has a 72.63\% overlap with the disambiguation bias challenge set in the OS18 domain and 64.4\% overlap in the WMT19 domain. Thus, disambiguation biases appear to indeed capture some sense frequency effects, which themselves represent a dataset artifact, but also introduce novel information.

Our experimental findings indicate that translation models leverage undesirable surface-level correlations when resolving lexical ambiguity and are prone to disambiguation errors in cases where learned statistical patterns are violated. Next, we use these insights for the construction of adversarial samples that cause disambiguation errors by minimally perturbing source sentences.

\section{Adversarial WSD attacks on NMT}
\label{sec:adv_wsd}
\begin{table*}[!t]
\centering
\resizebox{12cm}{!}{%
\begin{tabular}{l l}
\textbf{IH} & During this \textcolor{red}{\textbf{first}} \textcolor{blue}{\textbf{spring}}, he planted another tree that looked the same.\\
\midrule
\textbf{RH} & A \textcolor{red}{\textbf{hot}} \st{new} \textcolor{blue}{\textbf{spring}} will conquer the dark nights of winter.\\
\midrule
\textbf{InH} & Come the \textcolor{blue}{\textbf{spring}}, I will be invading the \textcolor{red}{\textbf{whole}} country called Frankia.\\
\midrule
\textbf{RnH} & After a long, \textcolor{red}{\textbf{eternal}} \st{fallow} winter, \textcolor{blue}{\textbf{spring}} has come again to Fredericks Manor.\\
\end{tabular}}
\caption{Perturbation examples; seed sense: \textit{season}, adversarial sense: \textit{water source}. Insertion/replacement in red.}
\label{tab:gen_ex}
\end{table*}

Adversarial attacks probe model robustness by attempting to elicit incorrect predictions with perturbed inputs \cite{zhang2020adversarial}. By crafting adversarial samples that explicitly target WSD capabilities of NMT models, we seek to provide further evidence for their susceptibility to dataset artifacts.

\subsection{Generating adversarial WSD samples}
\label{sec:gen_adv}
Our proposed attack strategy is based on the assumption that introducing an attractor into a sentence can flip its inherent disambiguation bias towards the attractor's sense cluster. Thus, translations of the so perturbed sentence will be more likely to contain WSD errors. The corresponding sample generation strategy consists of four stages: 
\begin{enumerate}
\item Select \textit{seed} sentences containing homographs to be adversarially perturbed. 
\item Identify attractors that are likely to yield fluent and natural samples.
\item Apply perturbations by introducing attractors into seed sentences.
\item Predict effective adversarial samples based on attractor properties. 
\end{enumerate}
The targeted attack is deemed successful if a victim model accurately translates the homograph in the seed sentence, but fails to correctly disambiguate it in the adversarially perturbed sample, instead translating it as one of the senses belonging to the attractor's sense cluster. This is a significantly more challenging attack success criterion than the general reduction in test BLEU typically employed for evaluating adversarial attacks on NMT systems \cite{cheng2019robust}. Samples are generated using homographs and attractors collected in section \ref{sec:quantifying}, while all test sentence pairs extracted in section \ref{sec:probing} form the domain-specific seed sentence pools. Attack success is evaluated on the same baseline translation models as used throughout section \ref{sec:can_wsd}.

\subsection*{Seed sentence selection}
In order to generate informative and interesting adversarial samples, we focus on seed sentences that are likely to be unambiguous. We thus apply three filtering heuristics to seed sentence pairs:
\begin{itemize}
\item Sentences have to be at least 10 tokens long.
\item We mask out the correct homograph sense in the reference translation and use a pre-trained German BERT model \cite{devlin2019bert}\footnote{We use the implementation provided by the Hugging Face Transformers library \cite{wolf2019transformers}. We do not fine-tune BERT, as our use case corresponds to its original masked language modeling objective.} to predict it. Pairs are rejected if the most probable sense does not belong to the correct sense cluster which suggests that the sentence context may be insufficient for correctly disambiguating the homograph. As a result, WSD errors observed in model-generated translations of the constructed adversarial samples are more likely to be due to the applied adversarial perturbations.
\item 10\% of pairs with the highest disambiguation bias towards incorrect sense clusters are removed from the seed pool.
\end{itemize}
Setting the rejection threshold above 10\% can further reduce WSD errors in seed sentences. At the same time, it would likely render minimal perturbations ineffective, due to the sentences’ strong bias towards the correct homograph sense. Thus, we aim for a working compromise.

\subsection*{Perturbation types}
Naively introducing new words into sentences is expected to yield disfluent, unnatural samples. To counteract this, we constrain candidate attractors to adjectives, since they can usually be placed in front of English nouns without violating grammatical constraints. We consider four perturbation types: 
\begin{itemize}
\item Insertion of the attractor adjective in front of the homograph (IH) 
\item Replacement of a seed adjective modifying the homograph (RH) 
\item Insertion of the attractor adjective in front of a non-homograph noun (InH)
\item Replacement of a seed adjective modifying a non-homograph noun (RnH)
\end{itemize}
Replacement strategies require seed sentences to contain adjectives, but can potentially have a greater impact on the sentence's disambiguation bias by replacing attractors belonging to the correct sense cluster. Examples for each generation strategy are given in Table \ref{tab:gen_ex}, with homographs highlighted in blue and added attractors in red.

\subsection*{Attractor selection}
Since adjectives are subject to selectional preferences of homograph senses, not every attractor will yield a semantically coherent adversarial sample. For instance, inserting the attractor \textit{flying} in front of the homograph \textit{bat} in a sentence about baseball will likely produce a nonsensical expression, whereas an attractor like \textit{huge} would be more acceptable. We attempt to control for this type of disfluency by only considering attractors that had been previously observed to modify the homograph in its seed sentence sense. For non-homograph perturbations, attractors must have been observed modifying the non-homograph noun. This is ensured by obtaining a dependency parse for each sentence in the English half of the training data and maintaining a list of modifier adjectives for each known target homograph sense set and source noun.\footnote{This assumes correctness of homograph reference translations, which is unfortunately not always guaranteed.}

Lastly, to facilitate the fluency and naturalness of adversarial samples, the generation process incorporates a series of constraints:
\begin{itemize}
\item Comparative and superlative adjective forms are excluded from the attractor pool.
\item Attractors may not modify compound nouns due to less transparent selectional preferences.
\item Attractors are not allowed next to other adjectives modifying the noun, to avoid violating the canonical English adjective order.
\end{itemize}
As all heuristics rely on POS taggers or dependency parsers,\footnote{We use spaCy in all cases.} they are not free of noise, occasionally yielding disfluent or unnatural samples.

We restrict the number of insertions or replacements to one, so as to maintain a high degree of semantic similarity between adversarial samples and seed sentences. A single seed sentence usually yields several samples, even after applying the aforementioned constraints. Importantly, we generate samples using all retained attractors at this stage, without selecting for expected attack success. 

\subsection*{Post-generation filtering}
To further ensure the naturalness of generated samples, sentence-level perplexity is computed for each seed sentence and adversarial sample using a pre-trained English GPT2 \cite{radford2019language} language model.\footnote{As implemented in the Transformers library.} Samples are rejected if their perplexity exceeds that of their corresponding seed sentence by more than 20\%. In total, we obtain a pool of $\sim$500K samples for the OS18 domain and $\sim$3.9M samples for the WMT19 domain. Each sample is translated by all in-domain models.

\subsection{Identifying effective attractors}
\label{sec:att_suc}
\begin{table*}[!h]
\centering
\resizebox{9.7cm}{!}{%
\begin{tabular}{l c c c c}
\toprule
\textbf{Model} & \textbf{FREQ\textsubscript{\xmark}} & \textbf{PPMI\textsubscript{\xmark}} & \textbf{FREQ\textsubscript{DIFF}} & \textbf{PPMI\textsubscript{DIFF}}\\
\midrule
\midrule
OS18 Transformer & 0.307 & 0.367 & \textbf{0.438} & 0.306\\ 
\midrule
OS18 LSTM & 0.258 & 0.261 & \textbf{0.375} & 0.227\\ 
\midrule
OS18 ConvS2S & 0.228 & 0.174 & \textbf{0.325} & 0.165\\ 
\midrule
\midrule
WMT19 Transformer& 0.241 & 0.241 & \textbf{0.264} & 0.224\\ 
\midrule
WMT19 LSTM & 0.278 & 0.256 & \textbf{0.316} & 0.231\\ 
\midrule
WMT19 ConvS2S & 0.304 & 0.270 & \textbf{0.328} & 0.216\\ 
\bottomrule
\end{tabular}}
\caption{Rank biserial correlation between attractors' disambiguation bias and attack success.}
\label{tab:adv_corr}
\end{table*}

The success of the proposed attack strategy relies on the selection of attractors that are highly likely to flip the homograph translation from the correct \textit{seed} sense towards an \textit{adversarial} sense belonging to the attractors' own sense set. To identify such attractors, we examine correlations between attractors' disambiguation biases and the effectiveness of adversarial samples containing them. The attractors' bias values are based either on co-occurrence frequencies (Eqn.\ref{eq:1}) or PPMI scores (Eqn.\ref{eq:2}) with the homographs' sense clusters. In particular, we examine the predictive power of an attractor's bias towards the adversarial sense cluster (DB\textsubscript{\xmark}) as well as the difference between its adversarial and seed bias values (DB\textsubscript{DIFF}). As before, RBC and MWU measures are used to estimate correlation strength, with Table \ref{tab:adv_corr} summarizing the results.

Similarly to the findings reported in section \ref{sec:probing}, all uncovered correlations are strong and statistically significant with p $<$ 1e-5 (see appendix \ref{appendix:effect_size} for effect size thresholds). Importantly, FREQ\textsubscript{DIFF} exhibits the strongest correlation in all cases. 

\begin{figure}[!h]
\centering
\includegraphics[width=0.8\linewidth]{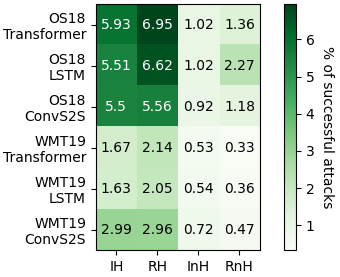}
\caption{Successful attacks per perturbation.}
\label{fig:gen_strats}
\end{figure}

We are furthermore interested in establishing which of the proposed perturbation methods yields most effective attacks. For this purpose, we examine the percentage of attack successes per perturbation strategy in Figure \ref{fig:gen_strats}, finding perturbations proximate to the homograph to be most effective.

\subsection*{Challenge set evaluation}
\begin{table*}[!t]
\centering
\resizebox{\linewidth}{!}{%
\begin{tabular}{l l l}
\toprule
\textbf{\underline{S}ource input / \underline{O}riginal output / \underline{P}erturbed output} & \textbf{Seed sense} & \textbf{Adv. sense}\\
\midrule
\textbf{S}: We played the songs again until we felt they sounded right, worked out all the (\textcolor{red}{\textbf{nasty}}) \textcolor{blue}{\textbf{bugs}}.\\\textbf{O}: Wir spielten die Lieder wieder, bis sie sich richtig anhörten und alle \textcolor{blue}{\textbf{Fehler}}\textsubscript{\cmark} ausarbeiteten.\\\textbf{P}: Wir spielten die Lieder wieder, bis sie sich richtig anhörten und alle \textcolor{red}{\textbf{bösen}} \textcolor{blue}{\textbf{Käfer}}\textsubscript{\xmark} ausarbeiteten. & \textit{error} & \textit{insect}\\
\midrule
\textbf{S}: The driver gets out, opens the (\textcolor{red}{\textbf{large}}) \textcolor{blue}{\textbf{boot}}, takes some flowers out to deliver.\\\textbf{O}: Der Fahrer steigt aus, öffnet den \textcolor{blue}{\textbf{Kofferraum}}\textsubscript{\cmark}, nimmt ein paar Blumen zum Ausliefern mit.\\\textbf{P}: Der Fahrer steigt aus, öffnet den \textcolor{red}{\textbf{großen}} \textcolor{blue}{\textbf{Stiefel}}\textsubscript{\xmark}, nimmt ein paar Blumen zum Ausliefern mit. & \textit{trunk} & \textit{shoe}\\
\midrule
\textbf{S}: The doctor somehow got that wig mixed up with the newspapers and (\textcolor{red}{\textbf{different}}) \textcolor{blue}{\textbf{letters}}.\\\textbf{O}: Der Arzt verwechselte die Perücke mit den Zeitungen und \textcolor{blue}{\textbf{Briefen}}\textsubscript{\cmark}.\\\textbf{P}: Der Arzt verwechselte die Perücke mit den Zeitungen und \textcolor{red}{\textbf{anderen}} \textcolor{blue}{\textbf{Buchstaben}}\textsubscript{\xmark}. & \textit{message} & \textit{character}\\
\midrule
\textbf{S}: And he will not cease until every \textcolor{red}{\textbf{last}} \textcolor{blue}{\textbf{race}} of the Four Lands is destroyed.\\\textbf{O}: Und er wird nicht aufgeben, bis jede \textcolor{blue}{\textbf{Rasse}}\textsubscript{\cmark} der Vier Länder ausgelöscht ist.\\\textbf{P}: Und er wird nicht aufhören, bis jedes \textcolor{red}{\textbf{letzte}} \textcolor{blue}{\textbf{Rennen}}\textsubscript{\xmark} der Vier Länder zerstört ist. & \textit{ethnic group} & \textit{contest}\\
\bottomrule
\end{tabular}}
\caption{Examples of successful attacks on the OS18 transformer. Homographs are blue, attractors are red.}
\label{tab:suceessful_attacks}
\end{table*}

\begin{figure}[!h]
\centering
\includegraphics[width=8cm]{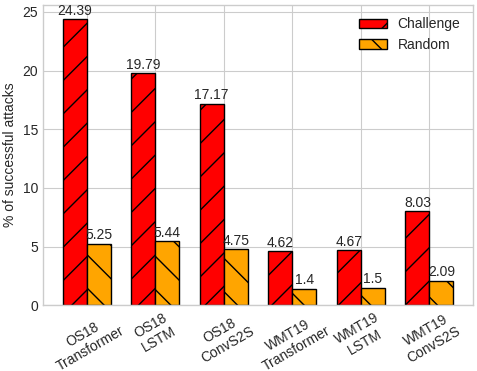}
\caption{Successful challenge sets attacks.}
\label{fig:adv_challenge}
\end{figure}

Having thus identified a strategy for selecting attractors that are likely to yield successful attacks, we construct a challenge set of 10000 adversarial samples with the highest attractor FREQ\textsubscript{DIFF} scores that had been obtained via the IH or RH perturbations. To enforce sample diversity, we limit the number of samples to at most 1000 per homograph. Additionally, we create equally-sized, secondary challenge sets by drawing samples at random from each domain's sample pool. Figure \ref{fig:adv_challenge} illustrates the attack success rate for both categories, while Table \ref{tab:suceessful_attacks} shows some of the successful attacks on the OS18 transformer. Further successful samples are reported in Appendix \ref{appendix:examples}.

The success rates are modest, ranging from 4.62\% to 24.39\%, but nonetheless showcase the capacity of targeted, minimal perturbations for flipping correct homograph translations towards a specific sense set. Since our attacks do not require access to model gradients or predictive score distributions, fall within the same domain as the models' training data, and have a strict notion of success, direct comparisons with previous work are difficult. Crucially, compared with a random sample selection strategy, subsampling informed by attractors' disambiguation bias is up to \textbf{4.25} times more successful at identifying effective adversarial samples.

While the relative improvement in attack success rate over the random baseline is comparable in both domains, the OS18 models are more susceptible to attacks in absolute terms. This may be due to their lower quality, or the properties of the training data, which can suffer from noisiness \cite{lison2019open}. Interestingly, the relative robustness of individual model architectures to WSD attacks also differs between domains, despite similar quality in terms of BLEU (see Table \ref{tab:baselines}). A more thorough investigation of architecture-specific WSD vulnerabilities is left for future work.

\subsection{Sample quality analysis}

To examine whether our adversarial samples would appear trivial and innocuous to human translators, automatic and human evaluation of samples included in the challenge set is conducted. Following \cite{morris2020reevaluating}, we use a grammar checker\footnote{\url{http://languagetool.org}} to evaluate the number of cases in which adversarial perturbations introduce grammatical errors. In the OS18 domain, only 1.04\% of samples are less grammatical than their respective seed sentences, whereas this is the case for 2.04\% of WMT19 samples, indicating a minimal degradation.

We additionally present two bilingual judges with 1000 samples picked at random from adversarial challenge sets in both domains and 1000 regular sentences from challenge sets constructed in section \ref{sec:probing}. For each adversarial source sentence, annotators were asked to choose whether the homograph's translation belongs to the correct or adversarial seed cluster. For each regular sentence, the choice was between the correct and randomly selected clusters. Across both domains, annotator error rate was 11.23\% in the adversarial setting and 11.45\% for regular sentences. As such, the generated samples display a similar degree of ambiguity to natural sentences that are likely to elicit WSD errors in NMT models. Annotator agreement was substantial (Cohen's kappa = 0.7). 

The same judges were also asked to rate the naturalness of each sentence on a Likert scale from 1 to 5. Perturbed sentences were assigned a mean score of 3.94, whereas regular sentences scored higher at 4.18. However, annotator agreement was low (weighted Kappa = 0.17). The observed drop in naturalness is likely due to the selection of attractors that are not fully consistent with the selectional preferences of homograph senses during sample generation. We attribute this to WSD errors in reference translations. For instance, we find that the attractor \textit{vampire} is occasionally applied to seed sentences containing the homograph \textit{bat} in its \textit{sporting equipment} sense, which can only occur if the attractor has been observed to modify this sense cluster in the training data (see \ref{sec:gen_adv}). Appendix \ref{appendix:instructions} replicates annotator instructions for both tasks.

\section{Transferability of adversarial samples}

An interesting question to consider is whether translation models trained on the same data are vulnerable to the same adversarial samples. We evaluate this by computing the Jaccard similarity index between successful attacks on each baseline model from the entire pool of adversarial samples described in section \ref{sec:att_suc}. We find the similarity to be low, raging between 10.1\% and 18.2\% for OS18 and between 5.7\% and 9.1\% for WMT19 samples, which suggests that different model architectures appear to be sensitive to different corpus artifacts, possibly due to differences in their inductive biases.

Considering the observed discrepancy in vulnerabilities between architectures, a natural follow-up question is whether two different instances of the same architecture are susceptible to the same set of attacks. We investigate this by training a second transformer model for each domain, keeping all settings constant with the initial models, but choosing a different seed for the random initialization. While the similarity between sets of successful adversarial samples is greater for two models of the same type, with 25.2\% in the OS18 and 12.4\% in WMT19 domain, is it still remarkably low.

\section{Literature review}

Polysemous terms represent a long-standing challenge for NMT. Past investigations sought to quantify the WSD capacity of translation models through challenge sets \cite{gonzales2017improving, raganato2019mucow}, to understand the disambiguation process by analysing their internal representations \cite{marvin2018exploring, tang2019encoders}, or to improve ambiguity resolution capabilities of translation models \cite{gonzales2017improving, liu2018handling}. To our knowledge, no study so far has examined the interaction between training data artifacts and WSD performance in detail.

Dataset artifacts, on the other hand, have previously been shown to enable models to make correct predictions based on incorrect or insufficient information \cite{mccoy2019right, gururangan2018annotation} by over-relying on spurious correlations present in the training data. Within NMT, models were found to exhibit gender-bias, reinforcing harmful stereotypes \cite{vanmassenhove2018getting, stanovsky2019evaluating}. As a response, strategies have been proposed for de-biasing the training data \cite{li2019repair, le2020adversarial}, as well as for making models more robust to data biases through adversarial training \cite{belinkov2019adversarial}. 

Adversarial attacks have recently been extended as an effective model analysis tool from vision to language tasks \cite{samanta2017towards, alzantot2018generating, glockner2018breaking, zhang2019generating}, including NMT \cite{cheng2018seq2sick, cheng2019robust}, where the focus so far has been on strategies requiring direct access to the victim model's loss gradient or output distribution. Recent surveys suggested that state-of-the-art attacks often yield ungrammatical and meaning-destroying samples, thus diminishing their usefulness for the evaluation of model robustness \cite{michel2019evaluation, morris2020reevaluating}. Targeted attacks on WSD abilities of translation models have so far remained unexplored.

\section{Conclusion}
We conducted an initial investigation into leveraging data artifacts for the prediction of WSD errors in machine translation and proposed a simple adversarial attack strategy based on the presented insights. Our results show that WSD is not yet a solved problem in NMT, and while the general performance of popular model architectures is high, we can identify or create sentences where models are more likely to fail due to data biases. 

The effectiveness of our methods owes to neural models struggling to accurately distinguish between meaningful lexical correlations and superficial ones. As such, the presented approach is expected to be transferable to other language pairs and translation directions, assuming that the employed translation models share this underlying weakness. Given the model-agnostic nature of our findings, this is likely to be the case.

As a continuation to this work, we intend to evaluate whether multilingual translation models are more resilient to lexical disambiguation biases and, as a consequence, are less susceptible to adversarial attacks that exploit source-side homography. Extending model-agnostic attack strategies to incorporate other types of dataset biases and to target natural language processing tasks other than machine translation is likewise a promising avenue for future research. Lastly, the targeted development of models that are resistant to dataset artifacts is a promising direction that is likely to aid generalization across linguistically diverse domains.

\section*{Acknowledgements}
We thank Sabine Weber and Tom Pelsmaeker for valuable discussions throughout the development of this work, as well as the anonymous reviewers for their constructive feedback. Rico Sennrich has received funding from the Swiss National Science Foundation (project MUTAMUR; no.\ 176727). 

\bibliographystyle{acl_natbib}
\bibliography{anthology,emnlp2020}

\begin{thebibliography}{42}
\expandafter\ifx\csname natexlab\endcsname\relax\def\natexlab#1{#1}\fi

\bibitem[{Alzantot et~al.(2018)Alzantot, Sharma, Elgohary, Ho, Srivastava, and
  Chang}]{alzantot2018generating}
Moustafa Alzantot, Yash~Sharma Sharma, Ahmed Elgohary, Bo-Jhang Ho, Mani
  Srivastava, and Kai-Wei Chang. 2018.
\newblock Generating natural language adversarial examples.
\newblock In \emph{Proceedings of the 2018 Conference on Empirical Methods in
  Natural Language Processing}.

\bibitem[{Barrault et~al.(2019)Barrault, Bojar, Costa-Juss{\`a}, Federmann,
  Fishel, Graham, Haddow, Huck, Koehn, Malmasi et~al.}]{barrault2019findings}
Lo{\"\i}c Barrault, Ond{\v{r}}ej Bojar, Marta~R Costa-Juss{\`a}, Christian
  Federmann, Mark Fishel, Yvette Graham, Barry Haddow, Matthias Huck, Philipp
  Koehn, Shervin Malmasi, et~al. 2019.
\newblock Findings of the 2019 conference on machine translation (wmt19).
\newblock In \emph{Proceedings of the Fourth Conference on Machine Translation
  (Volume 2: Shared Task Papers, Day 1)}, pages 1--61.

\bibitem[{Belinkov et~al.(2019)Belinkov, Poliak, Shieber, Van~Durme, and
  Rush}]{belinkov2019adversarial}
Yonatan Belinkov, Adam Poliak, Stuart Shieber, Benjamin Van~Durme, and
  Alexander~Sasha Rush. 2019.
\newblock On adversarial removal of hypothesis-only bias in natural language
  inference.
\newblock In \emph{Proceedings of the Joint Conference on Lexical and
  Computational Semantics}.

\bibitem[{Cheng et~al.(2018)Cheng, Yi, Zhang, Chen, and
  Hsieh}]{cheng2018seq2sick}
Minhao Cheng, Jinfeng Yi, Huan Zhang, Pin-Yu Chen, and Cho-Jui Hsieh. 2018.
\newblock Seq2sick: Evaluating the robustness of sequence-to-sequence models
  with adversarial examples.
\newblock \emph{arXiv preprint arXiv:1803.01128}.

\bibitem[{Cheng et~al.(2019)Cheng, Jiang, and Macherey}]{cheng2019robust}
Yong Cheng, Lu~Jiang, and Wolfgang Macherey. 2019.
\newblock Robust neural machine translation with doubly adversarial inputs.
\newblock In \emph{Proceedings of the 57th Annual Meeting of the Association
  for Computational Linguistics}, pages 4324--4333.

\bibitem[{Cohen(2013)}]{cohen2013statistical}
Jacob Cohen. 2013.
\newblock \emph{Statistical power analysis for the behavioral sciences}.
\newblock Academic press.

\bibitem[{Cureton(1956)}]{cureton1956rank}
Edward~E Cureton. 1956.
\newblock Rank-biserial correlation.
\newblock \emph{Psychometrika}, 21(3):287--290.

\bibitem[{Devlin et~al.(2019)Devlin, Chang, Lee, and
  Toutanova}]{devlin2019bert}
Jacob Devlin, Ming-Wei Chang, Kenton Lee, and Kristina Toutanova. 2019.
\newblock Bert: Pre-training of deep bidirectional transformers for language
  understanding.
\newblock In \emph{Proceedings of the 2019 Conference of the North American
  Chapter of the Association for Computational Linguistics: Human Language
  Technologies, Volume 1 (Long and Short Papers)}, pages 4171--4186.

\bibitem[{Dyer et~al.(2013)Dyer, Chahuneau, and Smith}]{dyer2013simple}
Chris Dyer, Victor Chahuneau, and Noah~A Smith. 2013.
\newblock A simple, fast, and effective reparameterization of ibm model 2.
\newblock In \emph{Proceedings of the 2013 Conference of the North American
  Chapter of the Association for Computational Linguistics: Human Language
  Technologies}, pages 644--648.

\bibitem[{Gehring et~al.(2017)Gehring, Auli, Grangier, and
  Dauphin}]{gehring2017convolutional}
Jonas Gehring, Michael Auli, David Grangier, and Yann Dauphin. 2017.
\newblock A convolutional encoder model for neural machine translation.
\newblock In \emph{Proceedings of the 55th Annual Meeting of the Association
  for Computational Linguistics (Volume 1: Long Papers)}, pages 123--135.

\bibitem[{Glockner et~al.(2018)Glockner, Shwartz, and
  Goldberg}]{glockner2018breaking}
Max Glockner, Vered Shwartz, and Yoav Goldberg. 2018.
\newblock Breaking nli systems with sentences that require simple lexical
  inferences.
\newblock In \emph{Proceedings of the 56th Annual Meeting of the Association
  for Computational Linguistics (Volume 2: Short Papers)}, pages 650--655.

\bibitem[{Gururangan et~al.(2018)Gururangan, Swayamdipta, Levy, Schwartz,
  Bowman, and Smith}]{gururangan2018annotation}
Suchin Gururangan, Swabha Swayamdipta, Omer Levy, Roy Schwartz, Samuel Bowman,
  and Noah~A Smith. 2018.
\newblock Annotation artifacts in natural language inference data.
\newblock In \emph{Proceedings of the 2018 Conference of the North American
  Chapter of the Association for Computational Linguistics: Human Language
  Technologies, Volume 2 (Short Papers)}, pages 107--112.

\bibitem[{Honnibal and Montani(2017)}]{honnibal2017spacy}
Matthew Honnibal and Ines Montani. 2017.
\newblock spacy 2: Natural language understanding with bloom embeddings,
  convolutional neural networks and incremental parsing.
\newblock \emph{To appear}, 7(1).

\bibitem[{Koehn et~al.(2007)Koehn, Hoang, Birch, Callison-Burch, Federico,
  Bertoldi, Cowan, Shen, Moran, Zens et~al.}]{koehn2007moses}
Philipp Koehn, Hieu Hoang, Alexandra Birch, Chris Callison-Burch, Marcello
  Federico, Nicola Bertoldi, Brooke Cowan, Wade Shen, Christine Moran, Richard
  Zens, et~al. 2007.
\newblock Moses: Open source toolkit for statistical machine translation.
\newblock In \emph{Proceedings of the 45th annual meeting of the association
  for computational linguistics companion volume proceedings of the demo and
  poster sessions}, pages 177--180.

\bibitem[{Le~Bras et~al.(2020)Le~Bras, Swayamdipta, Bhagavatula, Zellers,
  Peters, Sabharwal, and Choi}]{le2020adversarial}
Ronan Le~Bras, Swabha Swayamdipta, Chandra Bhagavatula, Rowan Zellers,
  Matthew~E Peters, Ashish Sabharwal, and Yejin Choi. 2020.
\newblock Adversarial filters of dataset biases.
\newblock \emph{arXiv}, pages arXiv--2002.

\bibitem[{Li and Vasconcelos(2019)}]{li2019repair}
Yi~Li and Nuno Vasconcelos. 2019.
\newblock Repair: Removing representation bias by dataset resampling.
\newblock In \emph{Proceedings of the IEEE Conference on Computer Vision and
  Pattern Recognition}, pages 9572--9581.

\bibitem[{Lison et~al.(2019)Lison, Tiedemann, Kouylekov et~al.}]{lison2019open}
Pierre Lison, J{\"o}rg Tiedemann, Milen Kouylekov, et~al. 2019.
\newblock Open subtitles 2018: Statistical rescoring of sentence alignments in
  large, noisy parallel corpora.
\newblock In \emph{LREC 2018, Eleventh International Conference on Language
  Resources and Evaluation}. European Language Resources Association (ELRA).

\bibitem[{Liu et~al.(2018)Liu, Lu, and Neubig}]{liu2018handling}
Frederick Liu, Han Lu, and Graham Neubig. 2018.
\newblock Handling homographs in neural machine translation.
\newblock In \emph{Proceedings of the 2018 Conference of the North American
  Chapter of the Association for Computational Linguistics: Human Language
  Technologies, Volume 1 (Long Papers)}, pages 1336--1345.

\bibitem[{Luong et~al.(2015)Luong, Pham, and Manning}]{luong2015effective}
Minh-Thang Luong, Hieu Pham, and Christopher~D Manning. 2015.
\newblock Effective approaches to attention-based neural machine translation.
\newblock In \emph{Proceedings of the 2015 Conference on Empirical Methods in
  Natural Language Processing}, pages 1412--1421.

\bibitem[{Mann and Whitney(1947)}]{mann1947test}
Henry~B Mann and Donald~R Whitney. 1947.
\newblock On a test of whether one of two random variables is stochastically
  larger than the other.
\newblock \emph{The annals of mathematical statistics}, pages 50--60.

\bibitem[{Marvin and Koehn(2018)}]{marvin2018exploring}
Rebecca Marvin and Philipp Koehn. 2018.
\newblock Exploring word sense disambiguation abilities of neural machine
  translation systems (non-archival extended abstract).
\newblock In \emph{Proceedings of the 13th Conference of the Association for
  Machine Translation in the Americas (Volume 1: Research Papers)}, pages
  125--131.

\bibitem[{McCoy et~al.(2019)McCoy, Pavlick, and Linzen}]{mccoy2019right}
Tom McCoy, Ellie Pavlick, and Tal Linzen. 2019.
\newblock Right for the wrong reasons: Diagnosing syntactic heuristics in
  natural language inference.
\newblock In \emph{Proceedings of the 57th Annual Meeting of the Association
  for Computational Linguistics}, pages 3428--3448.

\bibitem[{McGrath and Meyer(2006)}]{mcgrath2006effect}
Robert~E McGrath and Gregory~J Meyer. 2006.
\newblock When effect sizes disagree: the case of r and d.
\newblock \emph{Psychological methods}, 11(4):386.

\bibitem[{Michel et~al.(2019)Michel, Li, Neubig, and
  Pino}]{michel2019evaluation}
Paul Michel, Xian Li, Graham Neubig, and Juan~Miguel Pino. 2019.
\newblock On evaluation of adversarial perturbations for sequence-to-sequence
  models.
\newblock In \emph{Proceedings of NAACL-HLT}, pages 3103--3114.

\bibitem[{Morris et~al.(2020)Morris, Lifland, Lanchantin, Ji, and
  Qi}]{morris2020reevaluating}
John~X Morris, Eli Lifland, Jack Lanchantin, Yangfeng Ji, and Yanjun Qi. 2020.
\newblock Reevaluating adversarial examples in natural language.
\newblock \emph{arXiv preprint arXiv:2004.14174}.

\bibitem[{Navigli and Ponzetto(2010)}]{navigli2010babelnet}
Roberto Navigli and Simone~Paolo Ponzetto. 2010.
\newblock Babelnet: Building a very large multilingual semantic network.
\newblock In \emph{Proceedings of the 48th annual meeting of the association
  for computational linguistics}, pages 216--225. Association for Computational
  Linguistics.

\bibitem[{Ott et~al.(2019)Ott, Edunov, Baevski, Fan, Gross, Ng, Grangier, and
  Auli}]{ott2019fairseq}
Myle Ott, Sergey Edunov, Alexei Baevski, Angela Fan, Sam Gross, Nathan Ng,
  David Grangier, and Michael Auli. 2019.
\newblock fairseq: A fast, extensible toolkit for sequence modeling.
\newblock In \emph{Proceedings of NAACL-HLT 2019: Demonstrations}.

\bibitem[{Post(2018)}]{post2018call}
Matt Post. 2018.
\newblock A call for clarity in reporting bleu scores.
\newblock In \emph{Proceedings of the Third Conference on Machine Translation:
  Research Papers}, pages 186--191.

\bibitem[{Radford et~al.(2019)Radford, Wu, Child, Luan, Amodei, and
  Sutskever}]{radford2019language}
Alec Radford, Jeffrey Wu, Rewon Child, David Luan, Dario Amodei, and Ilya
  Sutskever. 2019.
\newblock Language models are unsupervised multitask learners.

\bibitem[{Raganato et~al.(2019)Raganato, Scherrer, and
  Tiedemann}]{raganato2019mucow}
Alessandro Raganato, Yves Scherrer, and J{\"o}rg Tiedemann. 2019.
\newblock The mucow test suite at wmt 2019: Automatically harvested
  multilingual contrastive word sense disambiguation test sets for machine
  translation.
\newblock In \emph{Proceedings of the Fourth Conference on Machine Translation
  (Volume 2: Shared Task Papers, Day 1)}, pages 470--480.

\bibitem[{Rios et~al.(2017)Rios, Mascarell, and
  Sennrich}]{gonzales2017improving}
Annette Rios, Laura Mascarell, and Rico Sennrich. 2017.
\newblock Improving word sense disambiguation in neural machine translation
  with sense embeddings.
\newblock In \emph{Proceedings of the Second Conference on Machine
  Translation}, pages 11--19.

\bibitem[{Ruscio(2008)}]{ruscio2008probability}
John Ruscio. 2008.
\newblock A probability-based measure of effect size: Robustness to base rates
  and other factors.
\newblock \emph{Psychological methods}, 13(1):19.

\bibitem[{Samanta and Mehta(2017)}]{samanta2017towards}
Suranjana Samanta and Sameep Mehta. 2017.
\newblock Towards crafting text adversarial samples.
\newblock \emph{arXiv preprint arXiv:1707.02812}.

\bibitem[{Sennrich et~al.(2016)Sennrich, Haddow, and
  Birch}]{sennrich2016neural}
Rico Sennrich, Barry Haddow, and Alexandra Birch. 2016.
\newblock Neural machine translation of rare words with subword units.
\newblock In \emph{Proceedings of the 54th Annual Meeting of the Association
  for Computational Linguistics (Volume 1: Long Papers)}, pages 1715--1725.

\bibitem[{Stanovsky et~al.(2019)Stanovsky, Smith, and
  Zettlemoyer}]{stanovsky2019evaluating}
Gabriel Stanovsky, Noah~A Smith, and Luke Zettlemoyer. 2019.
\newblock Evaluating gender bias in machine translation.
\newblock In \emph{Proceedings of the 57th Annual Meeting of the Association
  for Computational Linguistics}, pages 1679--1684.

\bibitem[{Tang et~al.(2019)Tang, Sennrich, and Nivre}]{tang2019encoders}
Gongbo Tang, Rico Sennrich, and Joakim Nivre. 2019.
\newblock Encoders help you disambiguate word senses in neural machine
  translation.
\newblock In \emph{Proceedings of the 2019 Conference on Empirical Methods in
  Natural Language Processing and the 9th International Joint Conference on
  Natural Language Processing (EMNLP-IJCNLP)}, pages 1429--1435.

\bibitem[{Vallat(2018)}]{Vallat2018}
Raphael Vallat. 2018.
\newblock Pingouin: statistics in python.
\newblock \emph{The Journal of Open Source Software}, 3(31):1026.

\bibitem[{Vanmassenhove et~al.(2018)Vanmassenhove, Hardmeier, and
  Way}]{vanmassenhove2018getting}
Eva Vanmassenhove, Christian Hardmeier, and Andy Way. 2018.
\newblock Getting gender right in neural machine translation.
\newblock In \emph{Proceedings of the 2018 Conference on Empirical Methods in
  Natural Language Processing}, pages 3003--3008.

\bibitem[{Vaswani et~al.(2017)Vaswani, Shazeer, Parmar, Uszkoreit, Jones,
  Gomez, Kaiser, and Polosukhin}]{vaswani2017attention}
Ashish Vaswani, Noam Shazeer, Niki Parmar, Jakob Uszkoreit, Llion Jones,
  Aidan~N Gomez, {\L}ukasz Kaiser, and Illia Polosukhin. 2017.
\newblock Attention is all you need.
\newblock In \emph{Advances in neural information processing systems}, pages
  5998--6008.

\bibitem[{Wolf et~al.(2019)Wolf, Debut, Sanh, Chaumond, Delangue, Moi, Cistac,
  Rault, Louf, Funtowicz et~al.}]{wolf2019transformers}
Thomas Wolf, Lysandre Debut, Victor Sanh, Julien Chaumond, Clement Delangue,
  Anthony Moi, Pierric Cistac, Tim Rault, R{\'e}mi Louf, Morgan Funtowicz,
  et~al. 2019.
\newblock Huggingface's transformers: State-of-the-art natural language
  processing.
\newblock \emph{arXiv preprint arXiv:1910.03771}.

\bibitem[{Zhang et~al.(2019)Zhang, Zhou, Miao, and Li}]{zhang2019generating}
Huangzhao Zhang, Hao Zhou, Ning Miao, and Lei Li. 2019.
\newblock Generating fluent adversarial examples for natural languages.
\newblock In \emph{Proceedings of the 57th Annual Meeting of the Association
  for Computational Linguistics}, pages 5564--5569.

\bibitem[{Zhang et~al.(2020)Zhang, Sheng, Alhazmi, and
  Li}]{zhang2020adversarial}
Wei~Emma Zhang, Quan~Z Sheng, Ahoud Alhazmi, and Chenliang Li. 2020.
\newblock Adversarial attacks on deep-learning models in natural language
  processing: A survey.
\newblock \emph{ACM Transactions on Intelligent Systems and Technology (TIST)},
  11(3):1--41.

\end{thebibliography}

\clearpage
\appendix
\section{Supplementary material}
\subsection{Data properties}
\label{appendix:data}
The WMT19 data is obtained by concatenating the Europarl v9, Common Crawl, and News Commentary v14 parallel corpora. Basic data cleaning is performed for both domains, which includes removal of pairs containing sentences classified by langid\footnote{\url{http://github.com/saffsd/langid.py}} as neither German or English and pairs with a source-to-target sentence length ratio exceeding 2. We create development and training splits for the OS18 domain by removing 10K sentence pairs from the full, shuffled corpus in each case. For each domain, we additionally hold-out 20\% of pairs to be used for the extraction of test pairs containing homographs, as described in section \ref{sec:probing}. Final statistics for the OS18 domain are reported in table \ref{tab:data_stats_os} and in \ref{tab:data_stats_wmt} for the WMT19 domain.

Each dataset is subsequently tokenized and truecased using Moses \cite{koehn2007moses} scripts\footnote{\url{http://github.com/moses-smt/mosesdecoder}}. For model training and evaluation, we additionally learn and apply BPE codes \cite{sennrich2016neural} to the data using the subword-NMT implementation\footnote{\url{http://github.com/rsennrich/subword-nmt}}, with 32k merge operations and the vocabulary threshold set to 50.

\subsection{Homograph list}
\label{appendix:homographs}
The full list of homographs used in our experiments is as follows: anchor, arm, band, bank, balance, bar, barrel, bark, bass, bat, battery, beam, board, bolt, boot, bow, brace, break, bug, butt, cabinet, capital, case, cast, chair, change, charge, chest, chip, clip, club, cock, counter, crane, cycle, date, deck, drill, drop, fall, fan, file, film, flat, fly, gum, hoe, hood, jam, jumper, lap, lead, letter, lock, mail, match, mine, mint, mold, mole, mortar, move, nail, note, offense, organ, pack, palm, pick, pitch, pitcher, plaster, plate, plot, pot, present, punch, quarter, race, racket, record, ruler, seal, sewer, scale, snare, spirit, spot, spring, staff, stock, subject, tank, tear, term, tie, toast, trunk, tube, vacuum, watch.

\subsection{Sense cluster examples}
Table \ref{tab:cluster_examples} lists some of the identified sense clusters for several homographs. All homographs used in our experiments have at least two sense clusters associated with them.

\subsection{Baseline models}
\label{appendix:models}
Table \ref{tab:models} provides implementation and training details for each architecture. Same settings are used for training identical models types in different domains. We use standard fairseq\footnote{\url{http://github.com/pytorch/fairseq}} \cite{ott2019fairseq} implementations for all model types and train them on NVIDIA 1080ti or NVIDIA 2080ti GPUs. Model translations are obtained by averaging the final 5 model checkpoints and decoding using beam search with beam size 5. 

\subsection{Base-rate adjusted effect size thresholds}
\label{appendix:effect_size}
Whether the effect size of correlations between dichotomous and quantitative variables can be considered strong depends on the size ratio between the two groups denoted by the dichotomous variable, i.e. its base rate. As the standard formulation of RBC is sensitive to the base rate, the estimated effect size decreases as the base rate becomes more extreme (see \cite{mcgrath2006effect} for details). Applied to our experimental setting, this means that the observed correlation values are sensitive to the number of sentences containing disambiguation errors relative to the amount of those that do not. This is an undesirable property, as we are only interested in the predictive power of our quantitative variables, regardless of how often disambiguation errors are observed. Thus, we adjust the thresholds for the interpretation of correlation strength to account for WSD errors being less frequent than WSD successes overall, in analogy to \cite{mcgrath2006effect}. Doing so enables the direct comparison of correlation strength between domains and model types, as each combination of the two factors exhibits a different disambiguation success base rate.

A common practice for interpreting effect size strength that does not account for base rate inequalities is the adoption of Cohen's benchmark \cite{cohen2013statistical}, which posits that the effect size $d$ is large if $d >= 0.8$, medium if $d >= 0.5$, and small if $d >= 0.2$. To adjust these threshold values for the observed base rates, they are converted according to Eqn. \ref{eq:8}, where $p1$ and $p2$ represent the proportions of groups described by the dichotomous variable, with $p_2 = 1 - p_1$:
\begin{equation}\label{eq:8}
threshold = \frac{d}{\sqrt{d^2 + \frac{1}{p_1, p_2}}}
\end{equation}
The adjusted effect size interpretation thresholds for WSD error correlation values as given in Table \ref{tab:seed_corr} are provided in Table \ref{tab:seed_effect}. Adjusted thresholds for attack success correlations as given in Table \ref{tab:adv_corr} are summarized in Table \ref{tab:adv_effect}.

\subsection{Annotator instructions}
\label{appendix:instructions}
The judges were presented with the following instructions for the described annotation tasks:

\textit{Your first task is to judge whether the meaning of the homograph as used in the given sentence is best described by the terms in the SENSE 1 cell or by those in the SENSE 2 cell. Please use the drop-down menu in the WHICH SENSE IS CORRECT? column to make your choice. If you think that neither sens captures the homograph's meaning, please select NONE from the options in the drop-down menu. If you think that the homograph as used in the given sentence can be equally interpreted both as SENSE 1 or SENSE 2, please select BOTH.}

\textit{We're also asking you to give us your subjective judgment whether the sentence you've been evaluating makes sense to you, i.e. whether it's grammatical, whether it can be easily understood, and whether it sounds acceptable to you as a whole. Typos and spelling mistakes, on the other hand, can be ignored. Specifically, we would like you to assign each sentence a naturalness score, ranging from 1 to 5, according to the following scale:}

\textit{\begin{itemize}
\item 1 = Completely unnatural (i.e. sentence is clearly ungrammatical, highly implausible, or meaningless / incoherent)
\item 2 = Somewhat unnatural (i.e. sentence is not outright incoherent, but sounds very strange)
\item 3 = Unsure (i.e. sentence is difficult to judge either way)
\item 4 = Mostly natural (i.e. sentence sounds good for the most part)
\item 5 = Completely natural (i.e. a well-formed English sentence)
\end{itemize}}

\textit{For instance a sentence like "John ate ten pancakes for breakfast." may get a ranking between 4 and 5, as it satisfies all of the above criteria. A sentence like "John ate green pancakes for breakfast." is grammatical but somewhat unusual and may therefore get a score between 3 and 4. "John ate late pancakes for breakfast.", on the other hand, does not sound very natural since pancakes cannot be "late" and may therefore be rated as 1 or 2. For this judgment we ask you to pay special attention to words in the neighborhood of the homograph. To submit your judgment please select the appropriate score from the drop-down menu in the DOES THE SENTENCE MAKE SENSE? column.}

\subsection{Examples of successful adversarial samples}
\label{appendix:examples}

Tables \ref{tab:suceessful_attacks_trans_OS18} - \ref{tab:suceessful_attacks_CNN_WMT19} list examples of successful adversarial attacks across the examined model architectures and dataset domains. As done throughout the paper, homographs are highlighted in blue, whereas the introduced attractors are emphasized in red.

\begin{table}[!h]
\centering
\resizebox{\linewidth}{!}{%
\begin{tabular}{l c c c}
\toprule
\textbf{Model} & \textbf{small} & \textbf{medium} & \textbf{large}\\
\midrule
\midrule
OS18 Transformer & 0.0542 & 0.1344 & 0.2121\\ 
\midrule
OS18 LSTM & 0.0666 & 0.1647 & 0.2581\\
\midrule
OS18 ConvS2S & 0.0710 & 0.1753 & 0.2740\\
\midrule
\midrule
WMT19 Transformer & 0.0381 & 0.0949 & 0.1508\\ 
\midrule
WMT19 LSTM & 0.0458 & 0.1138 & 0.1803\\
\midrule
WMT19 ConvS2S & 0.0502 & 0.1247 & 0.1971\\
\bottomrule
\end{tabular}}
\caption{Base-rate adjusted thresholds for the interpretation of WSD error prediction correlations.}
\label{tab:seed_effect}
\end{table}

\begin{table}[!h]
\centering
\resizebox{\linewidth}{!}{%
\begin{tabular}{l c c c}
\toprule
\textbf{Model} & \textbf{small} & \textbf{medium} & \textbf{large}\\
\midrule
\midrule
OS18 Transformer & 0.0339 & 0.0846 & 0.1345\\ 
\midrule
OS18 LSTM & 0.0338 & 0.0842 & 0.1340\\
\midrule
OS18 ConvS2S & 0.0328 & 0.0817 & 0.1301\\
\midrule
\midrule
WMT19 Transformer & 0.0166 & 0.0414 & 0.0661\\ 
\midrule
WMT19 LSTM & 0.0178 & 0.0446 & 0.0712\\
\midrule
WMT19 ConvS2S & 0.0219 & 0.0548 & 0.0874\\
\bottomrule
\end{tabular}}
\caption{Base-rate adjusted thresholds for the interpretation of attack success correlations.}
\label{tab:adv_effect}
\end{table}

\begin{table*}[!h]
\centering
\resizebox{12cm}{!}{%
\begin{tabular}{l c c c c}
\toprule
\textbf{Statistic} & \textbf{train} & \textbf{dev} & \textbf{test} & \textbf{held-out}\\
\midrule
\# sentences & 14,993,062 & 10,000 & 10,000 & 3,751,765\\
\midrule
\midrule
\# words (EN) & 106,873,835 & 71,719 & 71,332 & 26,763,351\\
\midrule
\# words/sentence (EN) & 7.13 & 7.17 & 7.13 & 7.13\\
\midrule
\midrule
\# words (DE) & 100,248,893 & 67,185 & 66,799 & 25,094,166\\
\midrule
\# words/sentence (DE) & 6.69 & 6,71 & 6.68 & 6.69\\
\bottomrule
\end{tabular}}
\caption{Corpus statistics for the OS18 domain.}
\label{tab:data_stats_os}
\end{table*}

\begin{table*}[!h]
\centering
\resizebox{14cm}{!}{%
\begin{tabular}{l c c c c c}
\toprule
\textbf{Statistic} & \textbf{train} & \textbf{dev (test18)} & \textbf{test14} & \textbf{test19} & \textbf{held-out} \\
\midrule
\# sentences & 4,861,743 & 2,998 & 3,003 & 1,997 & 1,215,435\\
\midrule
\midrule
\# words (EN) & 100,271,426 & 58,628 & 59,325 & 42034 & 25,057,036\\
\midrule
\# words/sentence (EN) & 20.62 & 19.56 & 19.76 & 21.05 & 20.62\\
\midrule
\midrule
\# words (DE) & 93,900,343 & 54,933 & 54,865 & 42,087 & 23,467,086\\
\midrule
\# words/sentence (DE) & 19.31 & 18.32 & 18.27 & 21.08 & 19.31\\
\bottomrule
\end{tabular}}
\caption{Corpus statistics for the WMT19 domain.}
\label{tab:data_stats_wmt}
\end{table*}

\label{appendix:clusters}
\begin{table*}[!h]
\centering
\resizebox{\textwidth}{!}{%
\begin{tabular}{c c c c}
\toprule
\textbf{Homograph} & \textbf{Sense 1} & \textbf{Sense 2} & \textbf{Sense 3}\\
\midrule
bat & \makecell{\textit{Chiroptera}, \textit{Fledertier},\\\textit{Handflügler}, \textit{Fledermaus},\\ \textit{Flattertier}} & \makecell{\textit{Schlagstock}, \textit{Schlagholz},\\ \textit{Baseballschläger}, \textit{Baseballkeule},\\\textit{Schläger}} & -\\
\midrule
case & \makecell{\textit{Karton}, \textit{Kiste},\\\textit{Päckchen}, \textit{Packung},\\\textit{Schachtel}, \textit{Kasten},\\\textit{Behälter}, \textit{Box}} & \makecell{\textit{Fall}, \textit{Zustand},\\\textit{Sache}, \textit{Gegebenheit},\\\textit{Lage}, \textit{Kontext},\\\textit{Umstand}, \textit{Status},\\\textit{Sachverhalt}, \textit{Stand},\\\textit{Situation}} &  \makecell{\textit{Prozess}, \textit{Gerichtsverfahren},\\\textit{Fall}, \textit{Gerichtsverhandlung},\\\textit{Sache}, \textit{Prozeß},\\\textit{Rechtsstreit}, \textit{Ermittlung},\\\textit{Antrag}, \textit{Rechtsfall},\\\textit{Gerichtsfall}, \textit{Klage},\\\textit{Verhör}, \textit{Rechtssache}}\\
\midrule
letter & \makecell{\textit{Sendschreiben}, \textit{Papierbrief},\\\textit{Musterbrief}, \textit{Anschreiben},\\\textit{Post}, \textit{Schreiben}, \textit{Brief}} & \makecell{\textit{Buchstabe}, \textit{Großbuchstabe},\\\textit{Charakter}, \textit{Letter},\\\textit{Kleinbuchstabe}, \textit{Zeichen}} & -\\
\midrule
spring & \makecell{\textit{Ringfeder}, \textit{Spiralfeder},\\\textit{Sprungfeder}, \textit{Feder},\\\textit{Tellerfeder}, \textit{Federung},\\\textit{Gummifeder}} & \makecell{\textit{Frühling}, \textit{Lenz},\\\textit{Frühjahr}} & \makecell{\textit{Quelle}, \textit{Brunnen},\\\textit{Quell}, \textit{Wasserquelle}}\\
\midrule
vacuum &  \makecell{\textit{Vakuum}, \textit{Nichts},\\\textit{Unterdruck}, \textit{Leerraum},\\\textit{Leere}, \textit{Luftleere}} & \makecell{\textit{Industriestaubsauger}, \textit{Staubsauger},\\\textit{Handstaubsauger}, \textit{Teppichkehrer},\\\textit{Bodenstaubsauger}, \textit{Allessauger},\\\textit{Sauger}, \textit{Kesselsauger}} & -\\
\bottomrule
\end{tabular}}
\caption{Non-exhaustive examples of homograph-specific sense clusters.}
\label{tab:cluster_examples}
\end{table*}

\begin{table*}[!h]
\centering
\resizebox{\textwidth}{!}{%
\begin{tabular}{l c c c}
\toprule
\textbf{Parameter} & \textbf{Transformer} & \textbf{LSTM} & \textbf{ConvS2S}\\
\midrule
batch size (subwords) & 24,576 & 4,096 & 4,096\\
\midrule
\# total updates & 100,000 & 600,000 & 600,000\\
\midrule
\# warm-up updates & 4,000 &  - & -\\
\midrule
\# updates between checkpoints & 1,000 & 4,000 & 4,000\\
\midrule
\# epochs between validations & 1 & 1 & 1\\
\midrule
optimizer & Adam & Adam & Adam\\
\midrule
Adam betas & 0.9, 0.98 & 0.9, 0.98 & 0.9, 0.98\\
\midrule
learning rate & scheduled (\textit{inverse\_sqrt}) & 0.0002 (+ decay) & 0.0003 (+ decay) \\
\midrule
\# total parameters (OS18) & 60,641,280 & 59,819,008 & 64,548,328\\
\midrule
\# total parameters (WMT19) & 61,714,432 & 60,892,160 & 66,696,728\\
\midrule
embedding size & 512 & 512 & 512\\
\midrule
Tied embeddings & Yes & Yes & Yes\\
\midrule
hidden size & 2,048 & 512 & 512\\
\midrule
\# encoder layers & 6 & 5 (bidirectional) & 8\\
\midrule
\# decoder layers & 6 & 5 & 8\\
\midrule
kernel size & - & - & 3 \\
\midrule
dropout & 0.1 & 0.2 & 0.2\\
\midrule
label smoothing & 0.1 & 0.1 & 0.1\\
\bottomrule
\end{tabular}}
\caption{Training settings and model hyperparameters.}
\label{tab:models}
\end{table*}

\begin{table*}[!h]
\centering
\resizebox{16cm}{!}{%
\begin{tabular}{l l l}
\toprule
\textbf{\underline{S}ource input / \underline{O}riginal output / \underline{P}erturbed output} & \textbf{Seed sense} & \textbf{Adv. sense}\\
\midrule
\textbf{S}: The Penguin was beating him with an (\textcolor{red}{\textbf{old}}) \textcolor{blue}{\textbf{bat}}, but it was Gordon that pulled the trigger.\\\textbf{O}: Der Pinguin hat ihn mit einem \textcolor{blue}{\textbf{Schläger}}\textsubscript{\cmark} geschlagen, aber Gordon hat abgedrückt.\\\textbf{P}: Der Pinguin hat ihn mit einer \textcolor{red}{\textbf{alten}} \textcolor{blue}{\textbf{Fledermaus}}\textsubscript{\xmark} geschlagen , aber Gordon hat abgedrückt. & \textit{club} & \textit{animal}\\
\midrule
\textbf{S}: I'm not going to relax until that thing its back in its (\textcolor{red}{\textbf{simple}}) \textcolor{blue}{\textbf{case}}.\\\textbf{O}: Ich werde mich nicht entspannen, bis dieses Ding nicht seinen Rücken in seinem\\\textcolor{blue}{\textbf{Koffer}}\textsubscript{\cmark} hat.\\\textbf{P}: Ich werde mich nicht entspannen, bis das Ding nicht seinen Rücken in seinem\\\textcolor{red}{\textbf{einfachen}} \textcolor{blue}{\textbf{Fall}}\textsubscript{\xmark} hat.\ & \textit{container} & \textit{instance} \\
\midrule
\textbf{S}: "They rest in their mother's (\textcolor{red}{\textbf{hot}}) \textcolor{blue}{\textbf{lap}}, enjoying the ultimate bliss"\\\textbf{O}: "Sie ruhen im \textcolor{blue}{\textbf{Schoß}}\textsubscript{\cmark} ihrer Mutter, genießen das ultimative Glück"\\\textbf{P}: "Sie ruhen in der \textcolor{red}{\textbf{heißen}} \textcolor{blue}{\textbf{Runde}}\textsubscript{\xmark} ihrer Mutter, genießen das ultimative Glück" & \textit{body part} & \textit{circuit}\\
\midrule
\textbf{S}: That's mighty neighbourly, but I got to play the (\textcolor{red}{\textbf{big}}) \textcolor{blue}{\textbf{organ}} for the parson tonight.\\\textbf{O}: Das ist mächtig nachbarschaftlich, aber ich muss heute Abend \textcolor{blue}{\textbf{Orgel}}\textsubscript{\cmark} für den\\Pfarrer spielen.\\\textbf{P}: Das ist mächtig nachbarschaftlich, aber ich muss heute Abend das \textcolor{blue}{\textbf{Organ}}\textsubscript{\xmark} für den\\Pfarrer spielen. & \textit{instrument} & \textit{body part}\\
\midrule
\textbf{S}: I'm just gonna write a (\textcolor{red}{\textbf{high}}) \textcolor{blue}{\textbf{note}}, and then we'll go.\\\textbf{O}: Ich schreibe nur einen \textcolor{blue}{\textbf{Zettel}}\textsubscript{\cmark} und dann gehen wir.\\\textbf{P}: Ich schreibe einen \textcolor{red}{\textbf{hohen}} \textcolor{blue}{\textbf{Ton}}\textsubscript{\xmark} und dann gehen wir. & \textit{writing} & \textit{tone}\\
\bottomrule
\end{tabular}}
\caption{Additional examples of successful attacks on the OS18 transformer. Homographs are blue, attractors are red.}
\label{tab:suceessful_attacks_trans_OS18}
\end{table*}

\begin{table*}[!h]
\centering
\resizebox{16cm}{!}{%
\begin{tabular}{l l l}
\toprule
\textbf{\underline{S}ource input / \underline{O}riginal output / \underline{P}erturbed output} & \textbf{Seed sense} & \textbf{Adv. sense}\\
\midrule
\textbf{S}: I only sell (\textcolor{red}{\textbf{good}}) \textcolor{blue}{\textbf{arms}} to people who fight clean wars! sure!\\\textbf{O}: Ich verkaufe nur \textcolor{blue}{\textbf{Waffen}}\textsubscript{\cmark} an Leute, die saubere Kriege bekämpfen.\\\textbf{P}: Ich verkaufe nur \textcolor{red}{\textbf{gute}} \textcolor{blue}{\textbf{Arme}}\textsubscript{\xmark} an Leute, die saubere Kriege bekämpfen. & \textit{weapon} & \textit{body part}\\
\midrule
\textbf{S}: We've heard they're trying to raise (\textcolor{red}{\textbf{new}}) \textcolor{blue}{\textbf{capital}} to rebuild their armies.\\\textbf{O}: Wir haben gehört, sie wollen \textcolor{blue}{\textbf{Kapital}}\textsubscript{\cmark} sammeln, um ihre Armeen aufzubauen.\\\textbf{P}: Wir haben gehört, dass sie eine \textcolor{red}{\textbf{neue}} \textcolor{blue}{\textbf{Hauptstadt}}\textsubscript{\xmark} aufziehen wollen, um ihre Armeen\\aufzubauen. & \textit{money} & \textit{city} \\
\midrule
\textbf{S}: Did you charge the Donellys for five (\textcolor{red}{\textbf{closed}}) \textcolor{blue}{\textbf{cases}} of vodka?\\\textbf{O}: Haben Sie die Donellys für fünf \textcolor{blue}{\textbf{Kisten}}\textsubscript{\cmark} Wodka berechnet?\\\textbf{P}: Haben Sie die Donellys für fünf \textcolor{red}{\textbf{geschlossene}} \textcolor{blue}{\textbf{Fälle}}\textsubscript{\xmark} Wodka berechnet? & \textit{container} & \textit{court case}\\
\midrule
\textbf{S}: All units, repeat. that is a battered yellow van, no (\textcolor{red}{\textbf{separate}}) \textcolor{blue}{\textbf{plates}}.\\\textbf{O}: An alle Einheiten, das ist ein gegrillter gelben Van, keine \textcolor{blue}{\textbf{Nummernschilder}}\textsubscript{\cmark}.\\\textbf{P}: An alle Einheiten, das ist ein gegrillter gelben Van, keine \textcolor{red}{\textbf{getrennten}} \textcolor{blue}{\textbf{Teller}}\textsubscript{\xmark}. & \textit{number plate} & \textit{dish}\\ 
\midrule
\textbf{S}: Um, (\textcolor{red}{\textbf{old}}) \textcolor{blue}{\textbf{seals}} tell the truth, but a sea lion's always lyin'?\\
\textbf{O}: \textcolor{blue}{\textbf{Robben}}\textsubscript{\cmark} sagen die Wahrheit, aber ein Seelöwen lügt immer ?\\
\textbf{P}: \textcolor{red}{\textbf{Alte}} \textcolor{blue}{\textbf{Siegel}}\textsubscript{\xmark} sagen die Wahrheit, aber ein Seelöwen lügt immer? & \textit{animal} & \textit{emblem}\\
\bottomrule
\end{tabular}}
\caption{Examples of successful attacks on the OS18 LSTM. Homographs are blue, attractors are red.}
\label{tab:suceessful_attacks_RNN_OS18}
\end{table*}

\begin{table*}[!h]
\centering
\resizebox{16cm}{!}{%
\begin{tabular}{l l l}
\toprule
\textbf{\underline{S}ource input / \underline{O}riginal output / \underline{P}erturbed output} & \textbf{Seed sense} & \textbf{Adv. sense}\\
\midrule
\textbf{S}: - Oh, well, keep the (\textcolor{red}{\textbf{small}}) \textcolor{blue}{\textbf{change}} and have a drink on me.\\\textbf{O}: Behalten Sie den \textcolor{blue}{\textbf{Rest}}\textsubscript{\cmark} und trinken Sie auf mich.\\\textbf{P}: Oh, nun, behalte die \textcolor{red}{\textbf{kleine}} \textcolor{blue}{\textbf{Veränderung}}\textsubscript{\xmark} und trink einen auf mich. & \textit{coins} & \textit{development}\\
\midrule
\textbf{S}: Do you know how that (\textcolor{red}{\textbf{specific}}) \textcolor{blue}{\textbf{date}} went, by any chance?\\\textbf{O}: Wissen Sie, wie das \textcolor{blue}{\textbf{Date}}\textsubscript{\cmark} gelaufen ist?\\\textbf{P}: Wissen Sie, wie das \textcolor{blue}{\textbf{Datum}}\textsubscript{\xmark} gelaufen ist? & \textit{meeting} & \textit{calendar date}\\
\midrule
\textbf{S}: Goal! (public address) An amazing last-minute third goal that takes Greenock\\into the (\textcolor{red}{\textbf{strong}}) \textcolor{blue}{\textbf{lead}}.\\\textbf{O}: Ein erstaunliches drittes drittes Ziel, das Greenock in die \textcolor{blue}{\textbf{Führung}}\textsubscript{\cmark} führt.\\\textbf{P}: Ein erstaunliches drittes Ziel, das Greenock in die \textcolor{red}{\textbf{starke}} \textcolor{blue}{\textbf{Spur}}\textsubscript{\xmark} führt. & \textit{first place} & \textit{clue} \\
\midrule
\textbf{S}: I mean, you seem like someone who plots out every (\textcolor{red}{\textbf{fucking}}) \textcolor{blue}{\textbf{move}}.\\\textbf{O}: Ich meine, Sie scheinen jemand zu sein, der jeden \textcolor{blue}{\textbf{Schritt}}\textsubscript{\cmark} aussticht.\\\textbf{P}: Ich meine, Sie scheinen jemand zu sein, der jede \textcolor{red}{\textbf{verdammte}} \textcolor{blue}{\textbf{Bewegung}}\textsubscript{\xmark}\\ausschüttet. & \textit{action} & \textit{movement}\\
\midrule
\textbf{S}: You know, if we get hungry, we eat some chips, have some (\textcolor{red}{\textbf{crazy}}) \textcolor{blue}{\textbf{punch}} ...\\\textbf{O}: Weißt du, wenn wir hungrig werden, essen wir ein paar Chips, haben etwas \textcolor{blue}{\textbf{Punsch}}\textsubscript{\cmark} ...\\\textbf{P}: Weißt du, wenn wir hungrig werden, essen wir ein paar Chips, haben einen\\\textcolor{red}{\textbf{verrückten}} \textcolor{blue}{\textbf{Schlag}}\textsubscript{\xmark} ... & \textit{drink} & \textit{hit}\\
\bottomrule
\end{tabular}}
\caption{Examples of successful attacks on the OS18 ConvS2S. Homographs are blue, attractors are red.}
\label{tab:suceessful_attacks_CNN_OS18}
\end{table*}

\begin{table*}[!h]
\centering
\resizebox{16cm}{!}{%
\begin{tabular}{l l l}
\toprule
\textbf{\underline{S}ource input / \underline{O}riginal output / \underline{P}erturbed output} & \textbf{Seed sense} & \textbf{Adv. sense}\\
\midrule
\textbf{S}: Copenhagen - Copenhagen, Denmark's (\textcolor{red}{\textbf{financial}}) \textcolor{blue}{\textbf{capital}}, wants to be the world's first\\CO2-neutral city by 2025.\\\textbf{O}: Kopenhagen - Kopenhagen, die \textcolor{blue}{\textbf{Hauptstadt}}\textsubscript{\cmark} Dänemarks, will bis 2025 die\\erste CO2-neutrale Stadt der Welt sein.\\\textbf{P}: Kopenhagen - Kopenhagen, das \textbf{\textcolor{red}{Finanz}\textcolor{blue}{kapital}}\textsubscript{\xmark} Dänemarks, will bis 2025 die erste\\CO2-neutrale Stadt der Welt sein. & \textit{city} & \textit{money}\\ 
\midrule
\textbf{S}: This is done by pricking the earlobe with a small lancet and taking a (\textcolor{red}{\textbf{real}}) \textcolor{blue}{\textbf{drop}} of blood.\\\textbf{O}: Dies geschieht, indem der Ohrwurm mit einem kleinen Lancet geprickt wird und ein\\\textcolor{blue}{\textbf{Tropfen}}\textsubscript{\cmark} Blut eingenommen wird.\\\textbf{P}: Dies geschieht, indem der Ohrwurm mit einem kleinen Lancet geprickt wird und ein\\\textcolor{red}{\textbf{richtiger}} Blut\textbf{\textcolor{blue}{abfall}}\textsubscript{\xmark} entsteht. & \textit{drop of liquid} & \textit{decrease}\\ 
\midrule
\textbf{S}: One (\textcolor{red}{\textbf{small}} \st{positive}) \textcolor{blue}{\textbf{note}} was from the Republic of Ireland, which saw its PMI grow to\\57.3, its highest level since the end of 1999.\\\textbf{O}: Eine positive \textcolor{blue}{\textbf{Anmerkung}}\textsubscript{\cmark} war die aus der Republik Irland, wo das PMI auf 57,3\\anstieg, das höchste Niveau seit Ende 1999.\\\textbf{P}: Ein \textcolor{red}{\textbf{kleiner}} \textcolor{blue}{\textbf{Schein}}\textsubscript{\xmark} stammt aus der Republik Irland, wo das PMI auf 57,3 anstieg,\\das höchste Niveau seit Ende 1999. & \textit{remark} & \textit{paper money}\\
\midrule
\textbf{S}: His epoch-making (\textcolor{red}{\textbf{full}}) \textcolor{blue}{\textbf{record}} "Free Jazz" was released by Atlantic Records at the\\dawn of that decade.\\\textbf{O}: Seine epochale \textcolor{blue}{\textbf{Platte}}\textsubscript{\cmark} "Free Jazz" wurde zu Beginn des Jahrzehnts von\\Atlantic Records veröffentlicht.\\\textbf{P}: Seine epochale \textcolor{blue}{\textbf{Aufzeichnung}}\textsubscript{\xmark} "Free Jazz" wurde zu Beginn des Jahrzehnts von\\Atlantic Records veröffentlicht. & \textit{musical medium} & \textit{document}\\ 
\midrule
\textbf{S}: After winter delivered an early dose of (\textcolor{red}{\textbf{natural}}) \textcolor{blue}{\textbf{spring}} last week, temperatures\\dropped again on Monday to a high of just 15.8C in the city.\\\textbf{O}: Nachdem der Winter vergangene Woche eine frühe \textcolor{blue}{\textbf{Frühjahr}}sdosis\textsubscript{\cmark}\\geliefert hatte, fielen die Temperaturen am Montag wieder auf einen Höchstwert von nur\\15,8C in der Stadt.\\\textbf{P}: Nachdem der Winter letzte Woche eine frühe Dosis \textbf{\textcolor{red}{Natur}\textcolor{blue}{quelle}}\textsubscript{\xmark} lieferte, fielen die\\Temperaturen am Montag wieder auf einen Höchstwert von nur 15,8C in der Stadt. & \textit{season} & \textit{water source}\\
\bottomrule
\end{tabular}}
\caption{Examples of successful attacks on the WMT19 transformer. Homographs are blue, attractors are red.}
\label{tab:suceessful_attacks_transformer_WMT19}
\end{table*}

\begin{table*}[!h]
\centering
\resizebox{16cm}{!}{%
\begin{tabular}{l l l}
\toprule
\textbf{\underline{S}ource input / \underline{O}riginal output / \underline{P}erturbed output} & \textbf{Seed sense} & \textbf{Adv. sense}\\
\midrule
\textbf{S}: A Thousand Splendid Suns is a story of two women's lives in Afghanistan, where women\\are equal, as a table or the (\textcolor{red}{\textbf{last}}) \textcolor{blue}{\textbf{chair}}.\\\textbf{O}: Ein Thousand Splendid Seine ist eine Geschichte von zwei Frauen in Afghanistan, wo\\Frauen gleich sind, als Tisch oder \textcolor{blue}{\textbf{Stuhl}}\textsubscript{\cmark}.\\\textbf{P}: Ein Thousand Splendid Seine ist eine Geschichte von zwei Frauen in Afghanistan, wo\\Frauen gleich sind, als Tisch oder als \textcolor{red}{\textbf{letzter}} \textcolor{blue}{\textbf{Vorsitzender}}\textsubscript{\xmark}. & \textit{furniture} & \textit{chairperson}\\ 
\midrule
\textbf{S}: See a (\textcolor{red}{\textbf{small}} \st{rapid}) \textcolor{blue}{\textbf{drop}} in your CO level once you stop smoking.\\\textbf{O}: Sehen Sie sich einen schnellen \textcolor{blue}{\textbf{Rückgang}}\textsubscript{\cmark} Ihrer CO-Ebene an, sobald Sie das\\Rauchen einstellen.\\\textbf{P}: Sehen Sie einen \textcolor{red}{\textbf{kleinen}} \textcolor{blue}{\textbf{Tropfen}}\textsubscript{\xmark} auf Ihrem CO-Niveau, sobald Sie aufhören,\\Rauchen zu beenden. & \textit{decrease} & \textit{drop of liquid}\\
\midrule
\textbf{S}: And moreover - each of our guests will get a (\textcolor{red}{\textbf{different}} \st{small}) \textcolor{blue}{\textbf{present}}!\\\textbf{O}: Und darüber hinaus wird jeder unserer Gäste ein kleines \textcolor{blue}{\textbf{Geschenk}}\textsubscript{\cmark} bekommen!\\\textbf{P}: Und darüber hinaus wird jeder unserer Gäste eine \textcolor{red}{\textbf{andere}} \textcolor{blue}{\textbf{Gegenwart}}\textsubscript{\xmark} bekommen! & \textit{gift} & \textit{current time} \\
\midrule
\textbf{S}: A (\textcolor{red}{\textbf{new}}) \textcolor{blue}{\textbf{record}} of every transaction made is kept, allowing for a complete audit if\\necessary.\\\textbf{O}: Ein \textcolor{blue}{\textbf{Datensatz}}\textsubscript{\cmark} jeder Transaktion wird gehalten, so dass erforderlichenfalls\\eine vollständige Prüfung möglich ist.\\\textbf{P}: Ein \textcolor{red}{\textbf{neuer}} \textcolor{blue}{\textbf{Rekord}}\textsubscript{\xmark} jeder Transaktion wird gehalten, so dass erforderlichenfalls\\eine vollständige Prüfung möglich ist. & \textit{document} & \textit{achievement}\\
\midrule
\textbf{S}: Britain's new trade deals with non-EU countries would also probably involve\\(\textcolor{red}{\textbf{political}} \st{worse}) \textcolor{blue}{\textbf{terms}}.\\\textbf{O}: Die neuen Handelsvereinbarungen Großbritanniens mit Nicht-EU-Ländern würden\\wahrscheinlich auch schlechtere \textcolor{blue}{\textbf{Bedingungen}}\textsubscript{\cmark} beinhalten.\\\textbf{P}: Großbritanniens neue Handelsabkommen mit Nicht-EU-Ländern würden\\wahrscheinlich auch \textcolor{red}{\textbf{politische}} \textcolor{blue}{\textbf{Begriffe}}\textsubscript{\xmark} beinhalten. & \textit{demand} & \textit{expression} \\
\bottomrule
\end{tabular}}
\caption{Examples of successful attacks on the WMT19 LSTM. Homographs are blue, attractors are red.}
\label{tab:suceessful_attacks_RNN_WMT19}
\end{table*}

\begin{table*}[!h]
\centering
\resizebox{16cm}{!}{%
\begin{tabular}{l l l}
\toprule
\textbf{\underline{S}ource input / \underline{O}riginal output / \underline{P}erturbed output} & \textbf{Seed sense} & \textbf{Adv. sense}\\
\midrule
\textbf{S}: Not to mention (non) uniform loading and soring fingers, contaminated with\\(\textcolor{red}{\textbf{common}}) \textcolor{blue}{\textbf{lead}}.\\\textbf{O}: Ganz zu schweigen von (nicht) einheitlichen Lade- und Sortierfingern, die mit\\\textcolor{blue}{\textbf{Blei}}\textsubscript{\cmark} kontaminiert sind.\\\textbf{P}: Ganz zu schweigen von (nicht) einheitlichen Lade- und Sortierfingern, die mit einer\\\textcolor{red}{\textbf{gemeinsamen}} \textcolor{blue}{\textbf{Führung}}\textsubscript{\xmark} kontaminiert sind. & \textit{metal} & \textit{first place}\\
\midrule
\textbf{S}: If the symbol "\&gt" is displayed, keep entering (\textcolor{red}{\textbf{greek}}) \textcolor{blue}{\textbf{letters}} until predictive options\\are displayed.\\\textbf{O}: Wenn das Symbol "\&gt" angezeigt wird, erhalten Sie die Eingabe von \textcolor{blue}{\textbf{Buchstaben}}\textsubscript{\cmark},\\bis prognostizierte Optionen angezeigt werden.\\\textbf{P}: Wenn das Symbol "\&gt" angezeigt wird, erhalten Sie immer wieder Gruß\textcolor{blue}{\textbf{briefe}}\textsubscript{\xmark},\\bis prognostizierte Optionen angezeigt werden. & \textit{character} & \textit{ message}\\
\midrule
\textbf{S}: This film is not about dialogue or a (\textcolor{red}{\textbf{little}} \st{stringent}) \textcolor{blue}{\textbf{plot}}, but all about atmosphere -\\a feverish dream that has become a film.\\\textbf{O}: In diesem Film geht es nicht um einen Dialog oder um eine strenge \textcolor{blue}{\textbf{Handlung}}\textsubscript{\cmark}, sondern\\um die Atmosphäre - ein feverser Traum, der zu einem Film geworden ist.\\\textbf{P}: In diesem Film geht es nicht um Dialog oder ein \textcolor{red}{\textbf{wenig}} \textcolor{blue}{\textbf{Grundstück}}\textsubscript{\xmark}, sondern alles über\\die Atmosphäre - ein feverser Traum, der zu einem Film geworden ist. & \textit{story} & \textit{tract of land}\\
\midrule
\textbf{S}: Manufacture of products from silicone and rubber, Production of springs,\\Manufacturing of springs, Winding of (\textcolor{red}{\textbf{small}}) \textcolor{blue}{\textbf{springs}}.\\\textbf{O}: Herstellung von Produkten aus Silikon- und Gummi, Herstellung von Quellen,\\Herstellung von Quellen, \textcolor{blue}{\textbf{Federn}}\textsubscript{\cmark}.\\\textbf{P}: Herstellung von Produkten aus Silikon- und Gummi, Herstellung von Quellen,\\Herstellung von Quellen, Winding von \textbf{\textcolor{red}{\textbf{kleinen}}} \textcolor{blue}{\textbf{Quellen}}\textsubscript{\xmark}. & \textit{device} & \textit{water source}\\ 
\midrule
\textbf{S}: In 1980, financial assets - (\textcolor{red}{\textbf{large}}) \textcolor{blue}{\textbf{stocks}}, bonds, and bank deposits - totaled around 100\%\\of GDP in the advanced economies.\\O; Im Jahr 1980 belief sich das Finanzvermögen - \textcolor{blue}{\textbf{Aktien}}\textsubscript{\cmark}, Anleihen und\\Bankeinlagen - in den hochentwickelten Volkswirtschaften rund 100\% des BIP.\\\textbf{P}: Im Jahr 1980 belief sich das Finanzvermögen - \textcolor{red}{\textbf{große}} \textcolor{blue}{\textbf{Bestände}}\textsubscript{\xmark}, Anleihen und\\Bankeinlagen - in den hochentwickelten Volkswirtschaften rund 100\% des BIP. & \textit{investment} & \textit{inventory}\\
\bottomrule
\end{tabular}}
\caption{Examples of successful attacks on the WMT19 ConvS2S. Homographs are blue, attractors are red.}
\label{tab:suceessful_attacks_CNN_WMT19}
\end{table*}
\end{document}